\documentclass[10pt, a4paper]{article}

\usepackage[final]{lrec-coling2024} 

\usepackage{booktabs}
\usepackage{stfloats}
\usepackage{subfig}
\usepackage{multirow}
\usepackage{soul}
\usepackage{titlesec}
\usepackage{amsthm,amsmath,amssymb}

\title{MoPE: Mixture of Prefix Experts for Zero-Shot Dialogue State Tracking}

\name{Tianwen Tang$^1$, Tong Zhu$^1$, Haodong Liu$^2$, Yin Bai$^2$, Jia Cheng$^2$, Wenliang Chen$^{1*}\thanks{*Corresponding Author}$} 

\address{$^1$School of Computer Science and Technology, Soochow University, China\\
         $^2$Meituan \\
         \{twtangtwtang, tzhu7\}@stu.suda.edu.cn, wlchen@suda.edu.cn \\
         \{liuhaodong05, baiyin, jia.cheng.sh\}@meituan.com \\
        }

\abstract{
Zero-shot dialogue state tracking (DST) transfers knowledge to unseen domains, reducing the cost of annotating new datasets.
Previous zero-shot DST models mainly suffer from domain transferring and partial prediction problems.
To address these challenges, we propose \textbf{M}ixture \textbf{o}f \textbf{P}refix \textbf{E}xperts (MoPE) to establish connections between similar slots in different domains, which strengthens the model transfer performance in unseen domains.
Empirical results demonstrate that MoPE-DST achieves the joint goal accuracy of 57.13\% on MultiWOZ2.1 and 55.40\% on SGD.\\ 
\newline 
 }

\begin{document}

\maketitleabstract

\section{Introduction}

Dialogue state tracking (DST) extracts and tracks the user's intention throughout a conversation in task-oriented dialogue (TOD) systems \citep{young2010hidden}.
The DST task is challenging due to the diversity and uncertainty of conversations, and it needs enormous data to train on a new domain.
Ideally, zero-shot DST could transfer knowledge to new domains, which reduces the efforts to build more datasets.
However, due to the large number of dialogue domains, there are two main challenges in zero-shot DST: 
(1) Domain transfer: 
It is impractical to collect dialogues involving all domains due to the infinite variety, so a DST model must have the capability to transfer to unseen domains.
(2) Partial-prediction:
DST models may predict fewer slot values when on a new domain.
This partial-prediction problem impedes TOD systems from providing accurate and necessary responses.


\begin{figure}
    \centering
    \includegraphics[width=.48\textwidth]{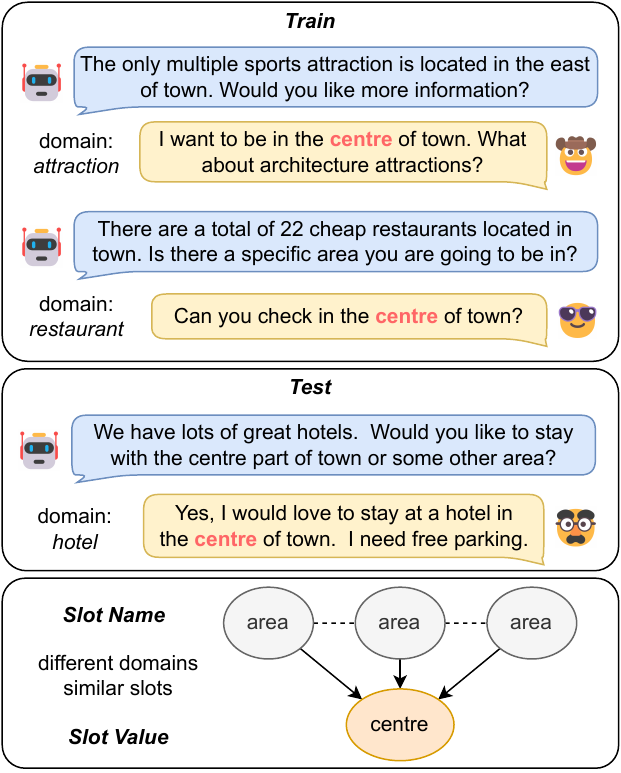}
    \caption{Illustration of dialogues in different domains share similar slot names even the same slot value.}
    \label{fig: dialogue}
\end{figure}

In order to transfer to unseen domains, \citet{wu-etal-2019-transferable} and \citet{heck-etal-2020-trippy} utilize the copy mechanism to generate slots.
However, these methods directly transfer to unseen domains without considering the differences with seen domains.
Consequently, the performance on unseen domains is notably lower than seen domains.
The primary reason for the low performance is that DST models need more relevant knowledge about unseen domains, and the information in dialogues involving these unseen domains is often neglected, leading to partial-prediction. 


To bridge the gap between seen and unseen domains, we explore the potential connections between them through similar slots.
As Figure~\ref{fig: dialogue} shows, We find that different domains may share some similar slots.
Even though the model is not trained on the hotel domain, the ``hotel area'' slot is similar to the trained slots ``attraction area'' and ``restaurant area'', and the model could refer to them when predicting on the unseen domain.
Based on the above considerations, we categorize all slots into different clusters and train a specialized expert for each cluster, which helps the slots from unseen domains find the most relevant expert.
Specialized experts can enhance the performance of slot prediction and reduce the occurrence of partial-prediction.

To address the above challenges and problems, we propose MoPE, which consists of a mixture of prefix experts on a pre-trained LLM.
We cluster similar slots with an unsupervised clustering algorithm and train a specialized expert for each cluster.
During inference, we utilize the cluster centroids to find the most relevant expert for the unseen slot and generate the corresponding dialogue state.
Considering the cost of training and the size of the whole model, we use the parameter-efficient fine-tuning (PEFT) method to train each expert where each expert is a specialized prefix prompt.

We conduct experiments on MultiWOZ2.1 and SGD datasets.
Experimental results demonstrate that our MoPE significantly outperforms all models with less than 10B parameters, achieving a remarkable 15\% increase in joint goal accuracy on both datasets.
Compared to large language models with extensive parameters like ChatGPT and Codex, MoPE achieves 0.20\% performance gain in joint goal accuracy on average.
From the clustering result, we observe that different domains can establish connections through similar slots.
Compared with sharing the same prefix prompt for all domains, using multiple specialized experts is helpful to the performance.

Our contributions are summarized as follows\footnote{ Our code is available at \href{https://github.com/ttw1018/MoPE-DST}{github.com/ttw1018/MoPE-DST}.}:

\begin{itemize}
    \item For the domain transfer in zero-shot DST, we establish connections between different domains through slots and apply multiple specialized experts to bridge the gap between seen domains and unseen domains.
    \item To reduce prediction errors and the training cost of multiple experts, we utilize a well-trained LLM and use prefix prompts as different experts to improve the condition generation of LLM with low training costs.
    \item We conduct experiments on two widely used dialogue state tracking benchmarks and achieve competitive performance, beating ChatGPT and Codex.
\end{itemize}

\section{Related Work}

\paragraph{Dialogue State Tracking}
DST plays a crucial role in natural language understanding within task-oriented dialogue systems. 
In the early years, DST methods \citep{lee-etal-2019-sumbt,zhang-etal-2020-find} heavily relied on manually crafted lexicons to capture dialogue states. 
However, this approach faced challenges in scaling up to longer and more intricate dialogues.
This difficulty arose from the need for more high-quality annotated data in emerging domains and the reliance on labor-intensive, hand-crafted lexicons.
To address these limitations, \citet{wu-etal-2019-transferable} and \citet{Le2020Non-Autoregressive} shifted their focus to open vocabulary DST research.
This transition aimed to diminish the reliance on manually crafted lexicons, offering a more adaptable and scalable approach.
With the widespread adoption of large language models, \citet{hu-etal-2022-context} and \citet{heck-etal-2023-chatgpt} have turned to powerful language models like Codex-Davinci-002 and ChatGPT to tackle the DST challenge.
However, these models have enormous parameters, making both training and inference processes difficult and costly.

Simultaneously, the approaches to solving the DST problem have become increasingly diverse.
\citet{gao2019dialog} reformulated DST as a reading comprehension task by answering the question: ``\textit{What is the state of the current dialogue?}"
\citet{shin-etal-2022-dialogue} framed DST as a dialogue summarization problem.
They trained a text-to-text template-based dialogue summary language model and recovered the dialogue state from the summarization using predefined rules.
\citet{hu-etal-2022-context} utilized a code-based large language model, formulating DST as a text-to-SQL problem, where the dialogue state is generated as an SQL query.

\paragraph{Parameter Efficient Transfer Learning for DST}
PETL for DST is designed to minimize the number of parameters requiring fine-tuning during domain transfer.
Despite tuning fewer parameters, several studies \citep{li-liang-2021-prefix, liu-etal-2022-p} have demonstrated that PETL can yield competitive results compared to traditional fine-tuning methods.
\citet{zhu-etal-2022-continual} introduced Continual Prompt Tuning, which prevents forgetting and facilitates knowledge transfer between tasks. This approach significantly enhances domain transfer capabilities.
\citet{aksu-etal-2023-prompter} employed prefix-tuning to customize models for new domains. They achieve this by utilizing descriptions of domain slots to generate dynamic prefix prompts.
However, these methods directly transfer the trained model to unseen domains, which often leads to a failure to establish connections between different domains.
MoE4DST \citep{wang-etal-2023-divide} partitions all observed data into semantically independent clusters and trains several adapters for each cluster. During inference, using a combination of adapters generates the dialogue state.
However, they cluster experts based on dialogue context rather than slot names, which leads to more granular connections between slots ignored, limiting the slot prediction's performance. Besides, the inconsistency of separate training and fusing inference is also a limitation of performance.
\section{Preliminary}

Dialogue State Tracking (DST) model aims at precisely predicting the dialogue state, where a dialogue state is represented as a triple in the form of domain-slot-value, such as (restaurant - food - Indian).
This prediction is based on both the dialogue history and predefined domains and slots.
Here, the domain signifies the dialogue topic, the slot is manually defined based on the domain, and the value is extracted from the  dialogue.
For ease of reference, throughout the remainder of this paper, we treat the ``slot" as a ``domain-slot" pair.

In this study, we approach the DST task as a question answering (QA) problem. 
The model utilizes the dialogue history as background knowledge and considers the predefined slot as the question. 
It then generates the dialogue state from the dialogue history, serving as the answer.

Formally, we define $D_t = [U_t, R_t]$ as a pair consisting of the system utterance $U_t$ and the user response $R_t$ in the $t$-th turn of the dialogue and $B_t$ represents the corresponding dialogue state.
$B_t$ is defined as a set of slot-value pairs, denoted as $B_t = \{(S_i, V_i) \mid i \in [1:N] \}$.
Here, $N$ represents the total number of dialogue states in the $t$-th turn, $S_i$ signifies the predefined slot pairs and $V_i$ corresponds to the slot value corresponding to $S_i$.
In summary, our approach involves providing the dialogue history $\{D_1 \cdots D_t \}$ and the predefined slot $S_i$, and then predicting the corresponding value $V_i$.

\section{Methodology}

\begin{figure*}
    \centering
    \includegraphics[width=.99\textwidth]{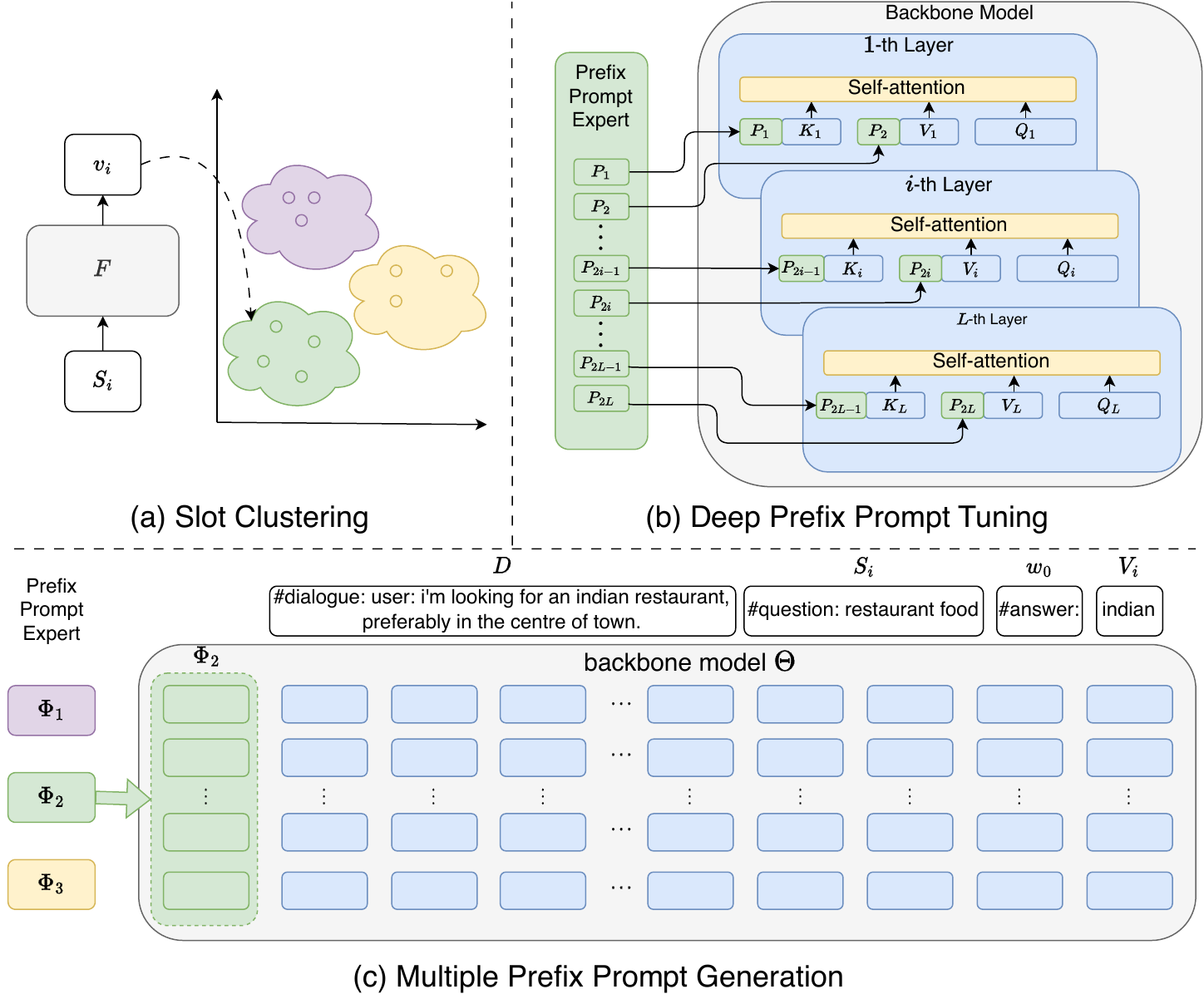}
    \caption{Illustration of our proposed method, including (a) Slot clustering, (b) Deep Prefix Prompt Tuning, and (c) Multiple Prefix Prompt Generation. Slot clustering is used to categorize all slots into distinct clusters and establishes connections between slots in different domains. Deep Prefix Prompt Tuning is our method to strengthen the LLM's conditional generation. Multiple Prefix Prompt Generation shows the complete pipeline of solving DST task.}
    \label{fig: structure}
\end{figure*}

Figure~\ref{fig: structure} provides an overview of our proposed method, which encompasses the following three key steps:

\begin{enumerate}
    \item First, we categorize all slots into distinct clusters using a clustering method.
    \item Next, we develop $K$ prefix prompt models for $K$ clusters, and prepare them for subsequent deep prefix prompt tuning.
    \item Lastly, we integrate the appropriate prefix prompt model for the slot into our backbone model, enabling the prediction of the corresponding value. Additionally, we optimize the prefix prompt model for enhanced performance.
\end{enumerate}

\subsection{Slot Clustering}

By dividing all slots into distinct clusters, similar slots are grouped together, allowing each cluster to predict values more accurately. This approach significantly improves precision when mixing all slots in a single cluster.
For instance, ``hotel area" and ``restaurant area" should be grouped in one cluster, considering their relevance to area information. 
On the other hand, ``hotel price range" and ``restaurant price range" should be grouped into another cluster since they both relate to price range information, which is distinct from area information. 
Clustering in this manner ensures that related slots are grouped, capturing the specific relationships between different types of information.
Slots within the same cluster exhibit similar semantic relations and have corresponding values in similar forms.
This shared similarity in both meaning and value format is beneficial for the value generation process.

Dividing slots into different clusters is a challenging task. 
In practical applications, manual slot clustering is daunting due to the large and increasing number of slots, coupled with the blurred and indistinguishable boundaries between slots. 
To address this issue, we utilize k-means clustering. 
We can use either the slot's feature or a combination of both slot and dialogue features for clustering.
However, considering the uncertainty associated with combining slot and dialogue features, we opt for using the slot's feature as the input for k-means in this study. 
This choice allows us to group similar slots into one cluster, ensuring each cluster is specialized and robust.

More specifically, given a slot $S_i$, we use a feature representation function $F$ to transform $S_i$ into a vector $v_i = F(S_i)$ within the semantic space. 
In this work, we explore two methods for feature representation: word embedding of the pre-trained language model and the hidden representation derived from the language model output.
Subsequently, we allocate each slot $S_i$ to one of the clusters $\mathcal{C}_k$ using the k-means algorithm based on $v_i$:

\begin{equation}
    \mathcal{C} _k = k \text{-} means( F(S_i) ), k \in \{1, \cdots ,K\}
\end{equation}
where $\mathcal{C} _k$ denotes the $k$-th cluster, and K represents the total number of clusters.

It is important to emphasize that all slots are predefined manually. 
Therefore, the k-means model should be initially fitted with the representations of all known slots. 
During inference, if there are unknown slots, their clusters $\mathcal{C} _k$ can be determined by the nearest cluster centroid labels.

\subsection{Deep Prefix Prompt Tuning}

To maximize the utilization of the pre-trained large language model and minimize resource consumption during training, we follow \citet{liu-etal-2022-p} to adopt parameter-efficient prefix prompts instead of fine-tuning the entire model.
The primary reason for avoiding fine-tuning is the necessity to train individual and specialized models for each cluster. Fine-tuning the entire model for each cluster demands substantial computing resources and leads to a linear increase in the overall model parameters with the number of clusters. Adopting an independent model approach mitigates these problems, leading to more efficient and manageable training processes.


In this work, we use deep prefix prompt tuning to train our autoregressive backbone model.
In detail, we incorporate $K$ additional prefix prompt models into the backbone model, each corresponding to one of the $K$ clusters. 
Each prefix prompt model $\Phi _k$ comprises $2L$ prefix prompts $P_i$, where $\Phi _k$ represents the prefix prompt model, $L$ is the number of layers of the backbone model, and $i \in \{1,\cdots,2L\}$. 
For the $l$-th layer of the backbone model, we concatenate $P_{2l-1}$ to the key of $l$ layer and $P_{2l}$ to the value of the $l$ layer:

\begin{equation}
    \begin{aligned}
        K_l & = [P_{2l-1}, K_l] \\
        V_l & = [P_{2l}, V_l]   \\
    \end{aligned}
\end{equation}
Where $l$ means the $l$-th layer of the backbone model, $K_l$ represents the key of the $l$-th layer, $V_l$ represents the value of the $l$-th layer, and $P_{k}$ is the $k$-th prefix prompt of the prefix prompt model.

Since we adopt a deep prefix prompt tuning approach, specifying a precise semantic prompt becomes challenging. Therefore, the prefix prompt is initialized randomly and subsequently trained with the data in the corresponding cluster. This method allows the model to adapt and learn the specific nuances of the cluster during the training process.

\subsection{Generation \& Optimization}


After dividing the slot $S_i$ into cluster $\mathcal{C}_k$ and obtaining the corresponding prefix prompt model $\Phi _k$, we concatenate the prefix prompt $p_k$ to the backbone model with the deep prefix prompting method. Subsequently, we generate the value $V_i$ in an autoregressive way:

\begin{equation}
    \begin{aligned}
        w_j = & argmax (p (w \mid D, S_i, w_0, \cdots, \\ & w_{j-1}; \Phi_k, \Theta)), j \in [1, L_{V_i}]
    \end{aligned}
\end{equation}

\begin{equation}
    V_i = \{ w_1, w_2, \cdots, w_{L_{V_i}} \}
\end{equation}
where $w_j$ is the $j$-th word in $V_i$, $D$ is the dialogue history, $S_i$ is the slot, $\Phi_k$ is the $k$-th prefix prompt model, $\Theta$ represents our backbone model, and $L_{V_i}$ is the length of $V_i$. 
Notably, given our QA approach to generating the dialogue state, $w_0$ is a predetermined word ``answer'' and indicates the answer context.

During training, we use teacher forcing to train the prefix prompts, and utilize cross-entropy loss to optimize the prefix prompts:
\begin{equation}
    \begin{aligned}
        \mathcal{L} = \sum_{j=1}^{L_{V_i}} - \hat{w_j}\log p(w_j \mid & D, S_i, w_0, \cdots, \\
                                                                      & w_{j-1}; \Phi_k, \Theta)
    \end{aligned}
\end{equation}
where $\mathcal{L}$ represents the loss of the model and $\hat{w_j}$ is $j$-th word of the ground truth $\hat{V_i}$ for the slot $S_i$.


Throughout the entire training process, we keep the parameters of the backbone model $\Theta$ fixed and only adjust the parameters of the prefix prompts $\Phi$ to minimize the loss $\mathcal{L}$.

\section{Experiments}

\subsection{Datasets}

We conduct experiments on two widely used DST datasets: MultiWOZ \citep{budzianowski-etal-2018-multiwoz} and SGD \citep{Rastogi_Zang_Sunkara_Gupta_Khaitan_2020}. 
MultiWOZ is a fully-labeled dataset consisting of human-human written conversations covering various domains and topics. 
It consists of over 8k dialogues spanning seven different domains and provides turn-level annotations and descriptions of each slot label. 
We use MultiWOZ version 2.1, which addresses the noisy state annotations in the original dataset \citep{eric-etal-2020-multiwoz}. 
To keep in line with previous studies, we limit our experiments to only five domains due to insufficient data for evaluation in the remaining two domains.
Similar to MultiWOZ, the SGD dataset is a fully labeled collection of machine-to-machine conversations across various domains and topics.
It comprises over 16K annotated conversations spanning over 20 diverse domains. 
Additionally, the dataset includes unseen domains in the test data, allowing for the evaluation of zero-shot performance, where models are tested on domains they have not been explicitly trained on.
More detailed information about MultiWOZ and SGD can be found in Table~\ref{table:datasets-info}.

\subsection{Baseline Models}

We compare our model with the following zero-shot DST methods.
\textbf{TRADE} \citep{wu-etal-2019-transferable} proposes a transferable dialogue state generator (TRADE) that uses a copy mechanism to generate dialogue states from utterances, which mitigates the reliance on domain ontology and strengthen the knowledge sharing across domains.
\textbf{SGD-baseline} \citep{Rastogi_Zang_Sunkara_Gupta_Khaitan_2020} encodes all the intents, slots, and slot values for categorical slots present in the schema into an embedded representation and uses a single model, shared among all domains, to make predictions.
\textbf{TransferQA} \citep{lin-etal-2021-zero} proposes a transferable generative QA model that reformulates DST as a QA task and uses a text-to-text model to extract dialogue states.
\textbf{IC-DST} \citep{hu-etal-2022-context} formulates DST as a text-to-SQL task and proposes an in-context learning (ICL) framework for DST, where a large language model (LLM) takes a test instance and a few exemplars as the input, and directly retrieve the dialogue state.
\textbf{ChatGPT} \citep{heck-etal-2023-chatgpt} presents preliminary experimental results on the ChatGPT research preview \citep{openai2021chatgpt} and evaluates the ability of ChatGPT as a dedicated and dynamic dialogue state tracker.
\textbf{Prompter} \citep{aksu-etal-2023-prompter} uses descriptions of target domain slots to generate dynamic prefixes and then trains adaptive prefixes with prefix-tuning for zero-shot DST.
\textbf{MoE4DST} \citep{wang-etal-2023-divide} partitions all observed data into semantically independent clusters and trains several adapters for each cluster.
During inference, using a combination of adapters generates the dialogue state. 

\subsection{Metrics}

\begin{table}
    \centering
    \begin{tabular}{ccc}
        \toprule
        \textbf{Metric} & \textbf{MultiWOZ} & \textbf{SGD} \\ 
        \midrule
        Language        & EN                & EN           \\ 
        Speakers        & H2H               & M2M          \\ 
        \#Domains       & 7                 & 16           \\ 
        \#Dialogues     & 8,438              & 16,142        \\ 
        \#Turns         & 115,424            & 329,964       \\ 
        Avg.domains     & 1.80              & 1.84         \\ 
        Avg.turns       & 13.7              & 20.4         \\ 
        \#Slots         & 25                & 214          \\ 
        \#Values        & 4,510              & 14,139        \\ 
        \bottomrule
    \end{tabular}
    \caption{Information of used task-oriented dialogue corpora. H2H represents human-to-human and M2M represents machine-to-machine. \# represents the number and Avg represents the average number of each dialogue.}
    \label{table:datasets-info}
\end{table}
\begin{table*}[htbp]
    \centering
    \resizebox{\textwidth}{!}{
        \begin{tabular}{cccccccc}
            \toprule
            \multirow{2}{*}{\textbf{Model}}           & \multirow{2}{*}{\textbf{Size}} & \multicolumn{6}{c}{\textbf{Joint Goal Accuray}}                                                                                      \\
                                                      &                                & Attraction                                      & Hotel          & Restaurant     & Taxi           & Train          & Average        \\
            \midrule
            TRADE \citep{wu-etal-2019-transferable}   & \multirow{3}{*}{<1B}           & 19.87                                           & 13.70          & 11.52          & 60.58          & 22.37          & 25.76          \\
            Prompter \citep{aksu-etal-2023-prompter}  &                                & 35.80                                           & 19.20          & 26.70          & 66.30          & 39.50          & 37.20          \\
            MoE4DST \citep{wang-etal-2023-divide}     &                                & 41.35                                           & 27.72          & 33.76          & 66.90          & 43.81          & 42.71          \\
            \midrule
            ChatGPT \citep{heck-etal-2023-chatgpt}    & \multirow{2}{*}{>100B}         & 52.70                                           & 42.00          & 55.80          & 70.90          & 60.80          & 56.44          \\
            IC-DST Codex \citep{hu-etal-2022-context} &                                & 59.97                                           & \textbf{46.69} & \textbf{57.28} & \textbf{71.35} & 49.37          & 56.93          \\
            \midrule
            \midrule
            Ours(DPPT)                                & \multirow{2}{*}{<10B}          & 56.99                                           & 31.37          & 52.44          & 70.63          & \textbf{63.97} & 55.08          \\
            Ours(MoPE)                                &                                & \textbf{60.39}                                  & 34.14          & 55.89          & 71.27          & \textbf{63.97} & \textbf{57.13} \\
            \bottomrule
        \end{tabular}
    }
    \caption{Zero-shot results on  MultiWOZ2.1. All results are reported in joint goal accuracy (\%) and the best results on each column are bolded. DPPT represents deep prefix prompt tuning and MoPE represents the mixture of prefix experts.}
    \label{table:zero-shot-result}
\end{table*}
\begin{table*}
    \centering
    \begin{tabular}{cccccc}
        \toprule
        \multirow{2}{*}{\textbf{Model}}                              & \multicolumn{5}{c}{\textbf{Joint Goal Accuray}}                                                                     \\
                                                                     & Alarm                                           & Messaging      & Payment        & Train          & Average        \\
        \midrule
        SGD-baseline \citep{Rastogi_Zang_Sunkara_Gupta_Khaitan_2020} & 57.70                                           & 10.20          & 11.50          & 13.50          & 20.50          \\
        TransferQA \citep{lin-etal-2021-zero}                        & 58.30                                           & 13.30          & 24.70          & 17.40          & 25.90          \\
        MoE4DST \citep{wang-etal-2023-divide}                        & 68.80                                           & 28.70          & 19.40          & 42.30          & 39.80          \\
        \midrule
        \midrule
        Ours (DPPT)                                                  & 81.52                                           & 59.93          & 30.45          & \textbf{46.32} & 54.55          \\
        Ours (MoPE)                                                  & \textbf{83.41}                                  & \textbf{60.56} & \textbf{31.33} & \textbf{46.32} & \textbf{55.40} \\
        \bottomrule
    \end{tabular}
    \caption{Zero-shot results on SGD. All results are reported in joint goal accuracy (\%) and the best results on each column are bolded.}
    \label{table: sgd-result}
\end{table*}

In the zero-shot experiments, we follow \citet{wu-etal-2019-transferable} to use slot accuracy(SA) and joint goal accuracy (JGA) as evaluation metrics. SA is used to measure the accuracy of individual slot predictions.
JGA evaluates the accuracy of slots for dialogue turns.
A turn is correct only if all values in the dialogue turn are predicted accurately. 
JGA provides a comprehensive measure of the model's ability to capture the entire context and generate correct predictions for all slots in a given dialogue turn.

\subsection{Settings}

Our model is implemented in PyTorch with transformers \citep{wolf-etal-2020-transformers}. During the slot dividing process, we utilize ChatGLM-6B \citep{du2022glm} as the slot feature representation model and then use k-means algorithm \citep{hartigan1979algorithm} from scikit-learn \citep{scikit-learn} as the clustering method. During the whole training, we freeze the parameters of the backbone model and optimize the prefix prompt model using AdamW \citep{loshchilov2018decoupled} with a learning rate set to 1e-2. The length of the prefix prompt is set to 10. It is worth noting that we train each prefix prompt model independently. For all experiments, we use one NVIDIA A100 (40G) GPU.


\subsection{Main Results}

\paragraph{Results on MultiWOZ} 
Table~\ref{table:zero-shot-result} shows the results of our proposed model on MultiWOZ under the zero-shot setting. 
We find that our MoPE outperforms all compared baselines on the average joint goal accuracy, achieving a 0.20\% performance gain over IC-DST Codex. Compared to DST models smaller than 10B parameters, our MoPE demonstrates an impressive improvement, achieving over 20\% increase in average joint goal accuracy. However, we find that the performance of our model in the hotel domain is notably lower compared to the other four domains.
This is probably because the hotel domain has many specialized slots, which share few similarities and correlations with other domains (i.e. ``hotel parking'', ``hotel stay''). We compute the specialized slot rate of five domains, ``hotel" and ``restaurant" contain 40\% and 28\% specialized slots respectively, while the remaining three domains do not have any specialized slots. 

\paragraph{Results on SGD} 
Table~\ref{table: sgd-result} shows the results of our proposed model on SGD. 
Our MoPE-DST significantly outperforms all compared baselines, with an increase of more than 15\% on average. 
We observe that our MoPE-DST performs exceptionally well on the ``alarm'' domain, achieving an impressive joint goal accuracy of 83.41\%. This exceptional performance can be attributed to the simplicity of the ``alarm'' domain, which only comprises two slots: ``alarm name'' and ``alarm time''.
Furthermore, there are many slots associated with ``name'' and ``time'' in seen domains, contributing to the model's accuracy in predicting these slots.

\begin{table*}[t]
    \centering
    \resizebox{\textwidth}{!}{
    \begin{tabular}{ccccccccccc}
        \toprule
        \multirow{2}{*}{\textbf{Prefix Prompt Type}} & \multicolumn{2}{c}{Attraction} & \multicolumn{2}{c}{Hotel} & \multicolumn{2}{c}{Restaurant} & \multicolumn{2}{c}{Taxi}                                 & \multicolumn{2}{c}{Train} \\
                                       & SA                             & JGA                       & SA                             & JGA                      & SA    & JGA   & SA    & JGA  &SA & JGA \\
        \midrule
        random prefix prompt      & 11.80                          & 49.14                     & 56.22                          & 0.44                     & 60.60 & 3.37  & 28.97 & 1.02 & 90.95 & 63.97 \\
        one prefix prompt (DPPT) & 81.83 & 56.99 & 82.36 & 31.37 & 89.98 & 52.44 & 88.06 & 70.63 & 90.95 & 63.97 \\
        specialized prefix prompt (MoPE)     & 83.28                         & 60.39                     & 84.06                          & 34.14                    & 90.87 & 55.89 & 87.75 & 71.27 & 90.95 & 63.97\\
        \bottomrule
    \end{tabular}
    }
    \caption{The slot accuracy (\%) and joint goal accuracy (\%) results of random prefix prompt, one prefix prompt, and specialized prefix prompt. For the random prefix prompt experiment, we assign a random expert for each cluster label.}
    \label{table: random-prompt}
\end{table*}

\begin{table}
    \centering
    \begin{tabular}{cccc}
        \toprule
        model                & num example & SA (w/o none) \\
        \midrule
        \multirow{4}{*}{ICL} & 0           & 6.32          \\
                             & 1           & 41.89         \\
                             & 3           & 45.17         \\
                             & 5           & 43.76         \\
        \midrule
        DPPT (5\%)           & 0           & 85.95         \\
        \bottomrule
    \end{tabular}

    \caption{The results of the in-context learning (ICL) and the deep prefix prompt tuning (DPPT) on MultiWOZ2.1. The results are reported in slot accuracy (\%). The result of DPPT is trained with 5\% training data and reported SA excludes the ``none" value when calculating.}

    \label{table: ICL}
\end{table}

\subsection{Comparison with ICL}

To thoroughly examine the initial capabilities of the LLM and assess the impact of deep prefix prompt tuning (DPPT) for LLM, we compare the results of in-context learning (ICL) and DPPT. 
We evaluate the performance of ICL and the performance of DPPT using limited training data, comprising approximately 5\% of the entire training data.
The experimental results presented in Table~\ref{table: ICL} indicate that the LLM struggles to effectively solve the DST problem with the initial frozen pre-trained model, even when up to 5 exemplars are used.
Compared to the frozen LLM, integrating a well-trained deep prefix prompt into the frozen LLM notably enhances its performance. 
This demonstrates that the pre-trained LLM might not be well trained on DST datasets, and a well-trained deep prefix prompt effectively enhances the LLM's ability in DST.


\subsection{Analysis on the specificity of prefix prompt}

To investigate the impact of prefix prompt specificity, we conduct experiments comparing the performance of random prefix prompt and the specialized prefix prompt.
As shown in Table~\ref{table: random-prompt}, using a random prefix prompt will cause a sharp drop in performance, indicating that well-trained prefix prompt is specialized for specific slots.

\subsection{Analysis on the feature of clustering}

\begin{figure}
    \centering
    \includegraphics[width=.45\textwidth]{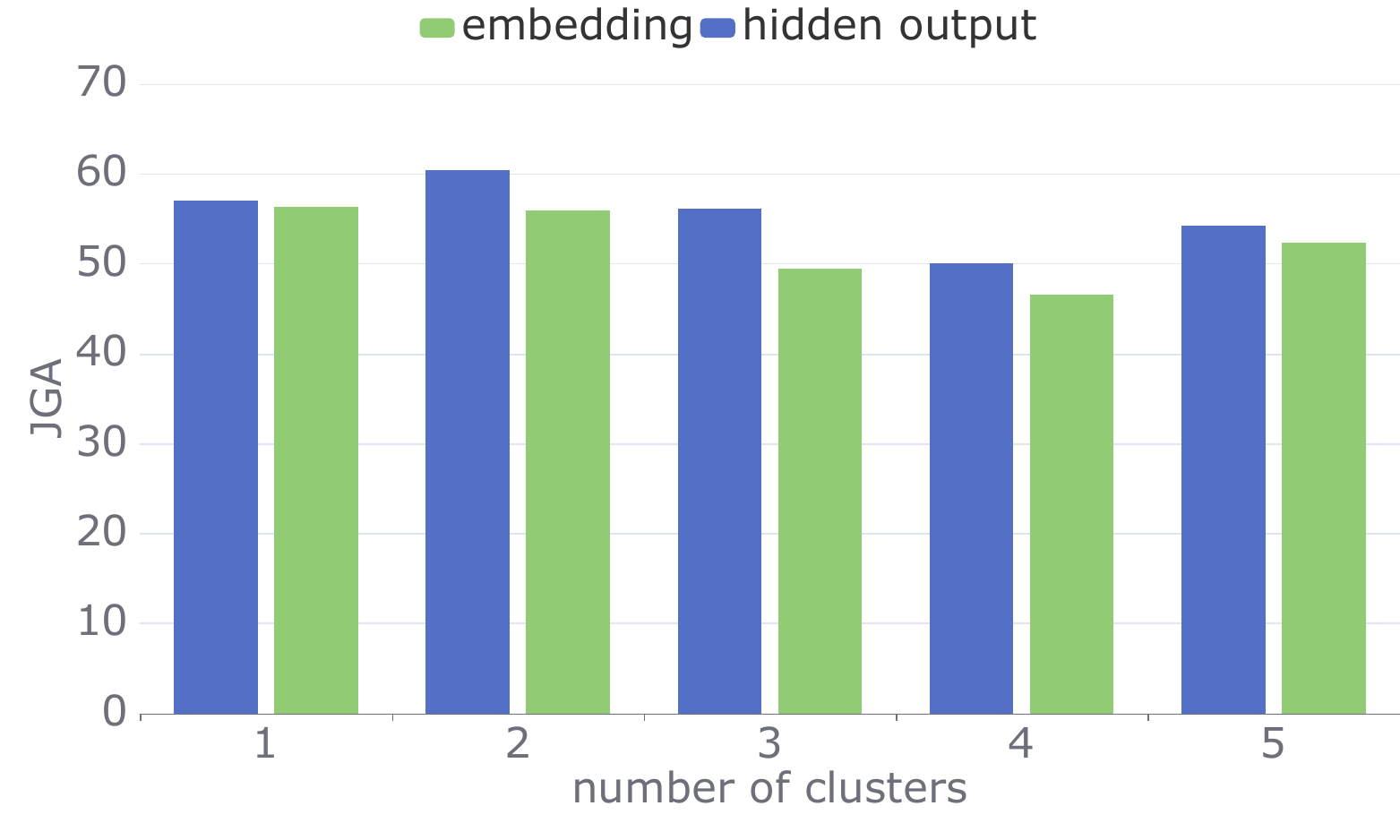}
    \caption{Zero-shot results on the attraction domain with different representations of slot feature.}
    \label{fig:compare-feature}
\end{figure}

The representation of the slot feature directly influences the clustering of slots and substantially impacts the final result. 
To study the effect of different feature representations of clustering on dialogue state tracking, we explore two ways for slot feature representation: the word embedding of the LLM and the hidden output of the LLM. 
The experimental results are shown in Figure~\ref{fig:compare-feature}.
We compare the results of different clustering features in the attraction domain under zero-shot experiments. 
We find that using hidden output as the clustering feature is significantly better than using the word embedding feature.
We find that clustering by hidden output can group more similar slots together.
Therefore, our subsequent experiments use the hidden output as the clustering feature.

\subsection{Analysis on the number of clusters}

To study the impact of the number of clusters, we conduct experiments to investigate the influence of the number of clusters. 
We compare the results of different numbers of clusters in five zero-shot domains.
As shown in Table~\ref{table: ACS}, as the number of clusters increases, the results tend to first improve and then decline.
The best results are achieved in three of five domains with 2 clusters.
For the remaining two domains, the best results are obtained with 1 and 3 clusters, respectively. 
We suspect that the diverse distribution of training slots influences the variation in the optimal cluster number.
Nevertheless, the results indicate that MoPE outperforms DPPT in most cases.

\subsection{Analysis on the similarity of slots}

\begin{table}
    \centering
    \begin{tabular}{ccccc}
        \toprule
        domain                      & K & train ACS       & test ACS        & JGA            \\
        \midrule


        \multirow{5}{*}{Attraction} & 1 & 0.6100          & \textbf{0.7718} & 56.99          \\
                                    & 2 & 0.6374          & \textbf{0.7718} & \textbf{60.39} \\
                                    & 3 & \textbf{0.6667} & \textbf{0.7718} & 56.11          \\
                                    & 4 & 0.6566          & \textbf{0.7718} & 50.02          \\
                                    & 5 & 0.6498          & 0.7650          & 54.27          \\

        \midrule


        \multirow{5}{*}{Hotel}      & 1 & 0.5907          & 0.7485          & 29.23          \\
                                    & 2 & 0.6041          & 0.7485          & 29.85          \\
                                    & 3 & 0.6680          & \textbf{0.7575} & \textbf{34.14} \\
                                    & 4 & \textbf{0.6682} & 0.7313          & 33.52          \\
                                    & 5 & 0.6031          & 0.7406          & 30.28          \\

        \midrule


        \multirow{5}{*}{Restaurant} & 1 & 0.6020          & \textbf{0.7319} & 52.44          \\
                                    & 2 & 0.6209          & \textbf{0.7319} & \textbf{55.89} \\
                                    & 3 & 0.6579          & \textbf{0.7319} & 47.53          \\
                                    & 4 & 0.6358          & 0.7044          & 50.26          \\
                                    & 5 & \textbf{0.6950} & \textbf{0.7319} & 44.92          \\

        \midrule

        \multirow{5}{*}{Taxi}       & 1 & 0.6310          & 0.6542          & 70.63          \\
                                    & 2 & 0.6508          & \textbf{0.8000} & \textbf{71.27} \\
                                    & 3 & 0.6849          & 0.6542          & 64.34          \\
                                    & 4 & 0.6851          & 0.6542          & 64.65          \\
                                    & 5 & \textbf{0.6876} & \textbf{0.8000} & 70.38          \\
        \midrule

        \multirow{5}{*}{Train}      & 1 & 0.6285          & \textbf{0.6440} & \textbf{63.97} \\
                                    & 2 & 0.6679          & 0.6175          & 55.12          \\
                                    & 3 & 0.6731          & 0.6363          & 46.11          \\
                                    & 4 & \textbf{0.7043} & 0.6174          & 43.86          \\
                                    & 5 & 0.6741          & 0.5768          & 50.84          \\
        \bottomrule
    \end{tabular}
    \caption{The average cosine similarity of slot feature. K represents the number of clusters of MoPE. ACS represents the average cosine similarity of slots. train ACS is the average cosine similarity of train slots and test ACS is the average cosine similarity of test slots.}
    \label{table: ACS}
\end{table}


We utilize the cosine similarity to measure the similarity between different slots and the average cosine similarity (ACS) of train and test slots is presented in Table~\ref{table: ACS}.
We find that a higher ACS of train slots does not ensure better performance, but better performance always aligns with higher ACS of test slots.

\begin{figure}
    \includegraphics[width=.47\textwidth]{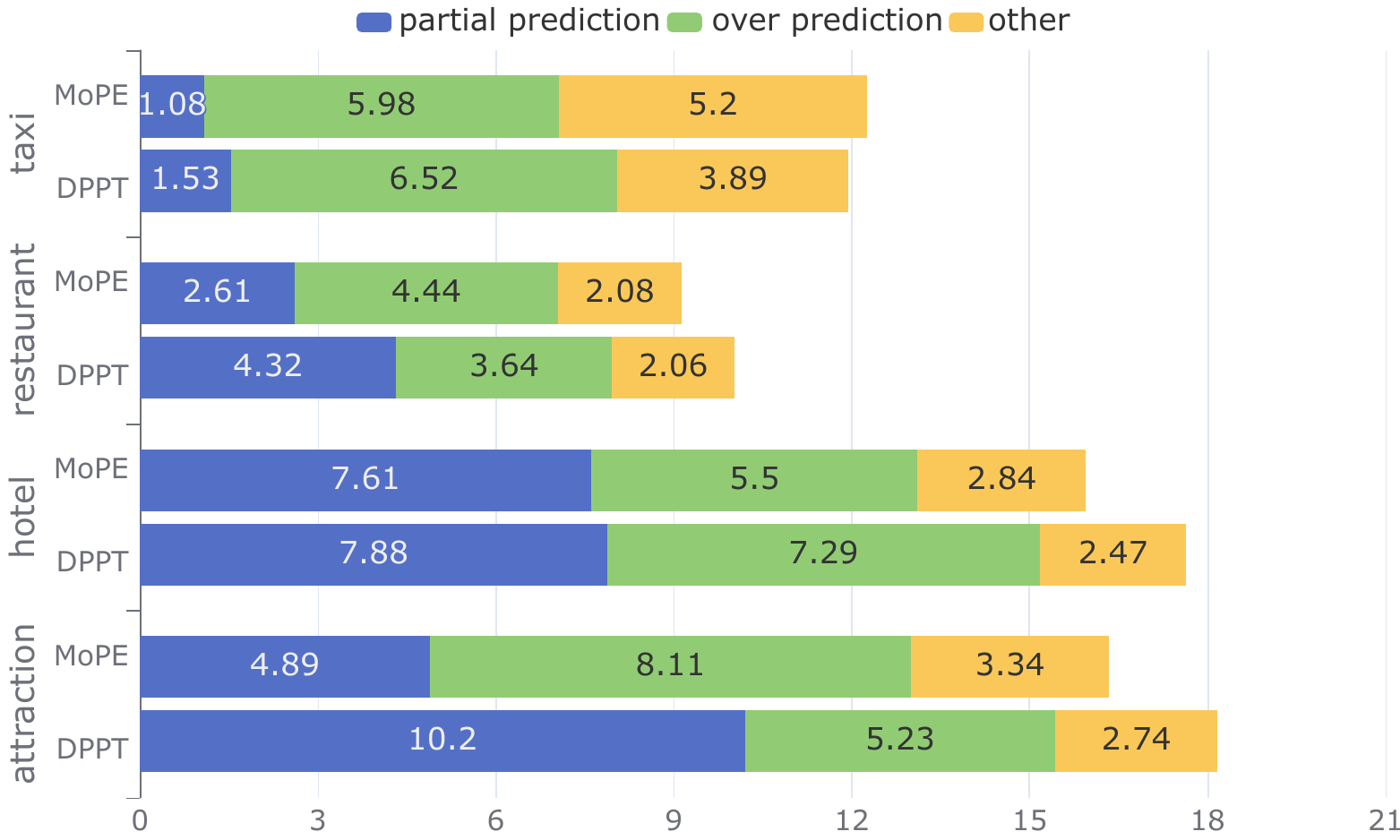}
    \caption{The slot error distribution of MoPE and DPPT.}
    \label{fig: error-rate}
\end{figure}

\subsection{Error Analysis}

The slot error distribution of MoPE and DPPT is presented in Figure~\ref{fig: error-rate}. 
We categorize slot errors into three types: partial-prediction, over-prediction, and other errors. 
``partial-prediction" and ``over-prediction" indicate model predicts less or more dialogue states, respectively. 
The slot error distribution shows that MoPE has fewer slot errors overall, especially in the category of ``partial-prediction''. 
We suspect that because specialized prompt makes LLM more sensitive to the relevant contents in the dialogue, which results in less ``partial-prediction". 
Therefore, the decrease in total slot errors leads to an overall improvement in performance.

\section{Conclusion}

In this paper, we propose a new method named MoPE to enhance the capability of LLM in solving the DST task.
The primary motivation behind this method is that the slots of different domains may share some common features and establishing the connections between slots from different domains is helpful to improve the performance for unseen domain prediction.
We categorize slots into different clusters and train a specialized expert for each cluster to improve the performance of unseen slots.
Moreover, we take a parameter-efficient fine-tuning approach to train specialized prefix prompts as experts, which significantly reduces the training cost.
Experimental results indicate that our method achieves competitive performances in zero-shot DST.

\section{Limitations}

We conclude the limitations of our method into two aspects. Firstly, our method benefits from different deep prefix prompts for different slots, which deeply depends on the way of dividing slots. In this work, we only use k-means as the clustering method and we believe that there are better clustering methods such as Brich \citep{zhang1996birch}, Agglomerative \citep{gowda1978agglomerative}, GMM \citep{yang2012robust}, and so on that can further improve the performance of the model. Secondly, our proposed method MoPE can be an independent part outside the model which means that can be a plug-in of LLMs to help them adapt to different tasks better, but we only experiment on the DST task with ChatGLM-6B.

\section*{Acknowledgements}


This work is supported by the National Natural Science Foundation of China (Grant No. 61936010 and 62036004). This work is also supported by Collaborative Innovation Center of Novel Software Technology and Industrialization, the Priority Academic Program Development of Jiangsu Higher Education Institutions, and the joint research project of Meituan and Soochow University.

\section*{References}
\bibliographystyle{lrec-coling2024-natbib}
\bibliography{cite}

\begin{thebibliography}{28}
\expandafter\ifx\csname natexlab\endcsname\relax\def\natexlab#1{#1}\fi

\bibitem[{Aksu et~al.(2023)Aksu, Kan, and Chen}]{aksu-etal-2023-prompter}
Ibrahim~Taha Aksu, Min-Yen Kan, and Nancy Chen. 2023.
\newblock \href {https://doi.org/10.18653/v1/2023.acl-long.252} {Prompter: Zero-shot adaptive prefixes for dialogue state tracking domain adaptation}.
\newblock In \emph{Proceedings of the 61st Annual Meeting of the Association for Computational Linguistics (Volume 1: Long Papers)}, pages 4588--4603, Toronto, Canada. Association for Computational Linguistics.

\bibitem[{Budzianowski et~al.(2018)Budzianowski, Wen, Tseng, Casanueva, Ultes, Ramadan, and Ga{\v{s}}i{\'c}}]{budzianowski-etal-2018-multiwoz}
Pawe{\l} Budzianowski, Tsung-Hsien Wen, Bo-Hsiang Tseng, I{\~n}igo Casanueva, Stefan Ultes, Osman Ramadan, and Milica Ga{\v{s}}i{\'c}. 2018.
\newblock \href {https://doi.org/10.18653/v1/D18-1547} {{M}ulti{WOZ} - a large-scale multi-domain {W}izard-of-{O}z dataset for task-oriented dialogue modelling}.
\newblock In \emph{Proceedings of the 2018 Conference on Empirical Methods in Natural Language Processing}, pages 5016--5026, Brussels, Belgium. Association for Computational Linguistics.

\bibitem[{Du et~al.(2022)Du, Qian, Liu, Ding, Qiu, Yang, and Tang}]{du2022glm}
Zhengxiao Du, Yujie Qian, Xiao Liu, Ming Ding, Jiezhong Qiu, Zhilin Yang, and Jie Tang. 2022.
\newblock Glm: General language model pretraining with autoregressive blank infilling.
\newblock In \emph{Proceedings of the 60th Annual Meeting of the Association for Computational Linguistics (Volume 1: Long Papers)}, pages 320--335.

\bibitem[{Eric et~al.(2020)Eric, Goel, Paul, Sethi, Agarwal, Gao, Kumar, Goyal, Ku, and Hakkani-Tur}]{eric-etal-2020-multiwoz}
Mihail Eric, Rahul Goel, Shachi Paul, Abhishek Sethi, Sanchit Agarwal, Shuyang Gao, Adarsh Kumar, Anuj Goyal, Peter Ku, and Dilek Hakkani-Tur. 2020.
\newblock \href {https://aclanthology.org/2020.lrec-1.53} {{M}ulti{WOZ} 2.1: A consolidated multi-domain dialogue dataset with state corrections and state tracking baselines}.
\newblock In \emph{Proceedings of the Twelfth Language Resources and Evaluation Conference}, pages 422--428, Marseille, France. European Language Resources Association.

\bibitem[{Gao et~al.(2019)Gao, Sethi, Agarwal, Chung, Hakkani-Tur, and AI}]{gao2019dialog}
Shuyang Gao, Abhishek Sethi, Sanchit Agarwal, Tagyoung Chung, Dilek Hakkani-Tur, and Amazon~Alexa AI. 2019.
\newblock Dialog state tracking: A neural reading comprehension approach.
\newblock In \emph{20th Annual Meeting of the Special Interest Group on Discourse and Dialogue}, page 264.

\bibitem[{Gowda and Krishna(1978)}]{gowda1978agglomerative}
K~Chidananda Gowda and GJPR Krishna. 1978.
\newblock Agglomerative clustering using the concept of mutual nearest neighbourhood.
\newblock \emph{Pattern recognition}, 10(2):105--112.

\bibitem[{Hartigan and Wong(1979)}]{hartigan1979algorithm}
John~A Hartigan and Manchek~A Wong. 1979.
\newblock Algorithm as 136: A k-means clustering algorithm.
\newblock \emph{Journal of the royal statistical society. series c (applied statistics)}, 28(1):100--108.

\bibitem[{Heck et~al.(2023)Heck, Lubis, Ruppik, Vukovic, Feng, Geishauser, Lin, van Niekerk, and Gasic}]{heck-etal-2023-chatgpt}
Michael Heck, Nurul Lubis, Benjamin Ruppik, Renato Vukovic, Shutong Feng, Christian Geishauser, Hsien-chin Lin, Carel van Niekerk, and Milica Gasic. 2023.
\newblock \href {https://doi.org/10.18653/v1/2023.acl-short.81} {{C}hat{GPT} for zero-shot dialogue state tracking: A solution or an opportunity?}
\newblock In \emph{Proceedings of the 61st Annual Meeting of the Association for Computational Linguistics (Volume 2: Short Papers)}, pages 936--950, Toronto, Canada. Association for Computational Linguistics.

\bibitem[{Heck et~al.(2020)Heck, van Niekerk, Lubis, Geishauser, Lin, Moresi, and Gasic}]{heck-etal-2020-trippy}
Michael Heck, Carel van Niekerk, Nurul Lubis, Christian Geishauser, Hsien-Chin Lin, Marco Moresi, and Milica Gasic. 2020.
\newblock \href {https://aclanthology.org/2020.sigdial-1.4} {{T}rip{P}y: A triple copy strategy for value independent neural dialog state tracking}.
\newblock In \emph{Proceedings of the 21th Annual Meeting of the Special Interest Group on Discourse and Dialogue}, pages 35--44, 1st virtual meeting. Association for Computational Linguistics.

\bibitem[{Hu et~al.(2022)Hu, Lee, Xie, Yu, Smith, and Ostendorf}]{hu-etal-2022-context}
Yushi Hu, Chia-Hsuan Lee, Tianbao Xie, Tao Yu, Noah~A. Smith, and Mari Ostendorf. 2022.
\newblock \href {https://doi.org/10.18653/v1/2022.findings-emnlp.193} {In-context learning for few-shot dialogue state tracking}.
\newblock In \emph{Findings of the Association for Computational Linguistics: EMNLP 2022}, pages 2627--2643, Abu Dhabi, United Arab Emirates. Association for Computational Linguistics.

\bibitem[{Le et~al.(2020)Le, Socher, and Hoi}]{Le2020Non-Autoregressive}
Hung Le, Richard Socher, and Steven~C.H. Hoi. 2020.
\newblock \href {https://openreview.net/forum?id=H1e_cC4twS} {Non-autoregressive dialog state tracking}.
\newblock In \emph{International Conference on Learning Representations}.

\bibitem[{Lee et~al.(2019)Lee, Lee, and Kim}]{lee-etal-2019-sumbt}
Hwaran Lee, Jinsik Lee, and Tae-Yoon Kim. 2019.
\newblock \href {https://doi.org/10.18653/v1/P19-1546} {{SUMBT}: Slot-utterance matching for universal and scalable belief tracking}.
\newblock In \emph{Proceedings of the 57th Annual Meeting of the Association for Computational Linguistics}, pages 5478--5483, Florence, Italy. Association for Computational Linguistics.

\bibitem[{Li and Liang(2021)}]{li-liang-2021-prefix}
Xiang~Lisa Li and Percy Liang. 2021.
\newblock \href {https://doi.org/10.18653/v1/2021.acl-long.353} {Prefix-tuning: Optimizing continuous prompts for generation}.
\newblock In \emph{Proceedings of the 59th Annual Meeting of the Association for Computational Linguistics and the 11th International Joint Conference on Natural Language Processing (Volume 1: Long Papers)}, pages 4582--4597, Online. Association for Computational Linguistics.

\bibitem[{Lin et~al.(2021)Lin, Liu, Madotto, Moon, Zhou, Crook, Wang, Yu, Cho, Subba, and Fung}]{lin-etal-2021-zero}
Zhaojiang Lin, Bing Liu, Andrea Madotto, Seungwhan Moon, Zhenpeng Zhou, Paul Crook, Zhiguang Wang, Zhou Yu, Eunjoon Cho, Rajen Subba, and Pascale Fung. 2021.
\newblock \href {https://doi.org/10.18653/v1/2021.emnlp-main.622} {Zero-shot dialogue state tracking via cross-task transfer}.
\newblock In \emph{Proceedings of the 2021 Conference on Empirical Methods in Natural Language Processing}, pages 7890--7900, Online and Punta Cana, Dominican Republic. Association for Computational Linguistics.

\bibitem[{Liu et~al.(2022)Liu, Ji, Fu, Tam, Du, Yang, and Tang}]{liu-etal-2022-p}
Xiao Liu, Kaixuan Ji, Yicheng Fu, Weng Tam, Zhengxiao Du, Zhilin Yang, and Jie Tang. 2022.
\newblock \href {https://doi.org/10.18653/v1/2022.acl-short.8} {{P}-tuning: Prompt tuning can be comparable to fine-tuning across scales and tasks}.
\newblock In \emph{Proceedings of the 60th Annual Meeting of the Association for Computational Linguistics (Volume 2: Short Papers)}, pages 61--68, Dublin, Ireland. Association for Computational Linguistics.

\bibitem[{Loshchilov and Hutter(2019)}]{loshchilov2018decoupled}
Ilya Loshchilov and Frank Hutter. 2019.
\newblock \href {https://openreview.net/forum?id=Bkg6RiCqY7} {Decoupled weight decay regularization}.
\newblock In \emph{International Conference on Learning Representations}.

\bibitem[{OpenAI(2021)}]{openai2021chatgpt}
OpenAI. 2021.
\newblock Chatgpt.
\newblock \url{https://www.openai.com/research/chatgpt/}.
\newblock Accessed: 2023-01-13.

\bibitem[{Pedregosa et~al.(2011)Pedregosa, Varoquaux, Gramfort, Michel, Thirion, Grisel, Blondel, Prettenhofer, Weiss, Dubourg, Vanderplas, Passos, Cournapeau, Brucher, Perrot, and Duchesnay}]{scikit-learn}
F.~Pedregosa, G.~Varoquaux, A.~Gramfort, V.~Michel, B.~Thirion, O.~Grisel, M.~Blondel, P.~Prettenhofer, R.~Weiss, V.~Dubourg, J.~Vanderplas, A.~Passos, D.~Cournapeau, M.~Brucher, M.~Perrot, and E.~Duchesnay. 2011.
\newblock Scikit-learn: Machine learning in {P}ython.
\newblock \emph{Journal of Machine Learning Research}, 12:2825--2830.

\bibitem[{Rastogi et~al.(2020)Rastogi, Zang, Sunkara, Gupta, and Khaitan}]{Rastogi_Zang_Sunkara_Gupta_Khaitan_2020}
Abhinav Rastogi, Xiaoxue Zang, Srinivas Sunkara, Raghav Gupta, and Pranav Khaitan. 2020.
\newblock \href {https://doi.org/10.1609/aaai.v34i05.6394} {Towards scalable multi-domain conversational agents: The schema-guided dialogue dataset}.
\newblock \emph{Proceedings of the AAAI Conference on Artificial Intelligence}, 34(05):8689--8696.

\bibitem[{Shin et~al.(2022)Shin, Yu, Moon, Madotto, and Park}]{shin-etal-2022-dialogue}
Jamin Shin, Hangyeol Yu, Hyeongdon Moon, Andrea Madotto, and Juneyoung Park. 2022.
\newblock \href {https://doi.org/10.18653/v1/2022.findings-acl.302} {Dialogue summaries as dialogue states ({DS}2), template-guided summarization for few-shot dialogue state tracking}.
\newblock In \emph{Findings of the Association for Computational Linguistics: ACL 2022}, pages 3824--3846, Dublin, Ireland. Association for Computational Linguistics.

\bibitem[{Wang et~al.(2023)Wang, Ding, Cao, Zhan, Lin, Wang, Tao, and Guo}]{wang-etal-2023-divide}
Qingyue Wang, Liang Ding, Yanan Cao, Yibing Zhan, Zheng Lin, Shi Wang, Dacheng Tao, and Li~Guo. 2023.
\newblock \href {https://doi.org/10.18653/v1/2023.acl-long.114} {Divide, conquer, and combine: Mixture of semantic-independent experts for zero-shot dialogue state tracking}.
\newblock In \emph{Proceedings of the 61st Annual Meeting of the Association for Computational Linguistics (Volume 1: Long Papers)}, pages 2048--2061, Toronto, Canada. Association for Computational Linguistics.

\bibitem[{Wolf et~al.(2020)Wolf, Debut, Sanh, Chaumond, Delangue, Moi, Cistac, Rault, Louf, Funtowicz, Davison, Shleifer, von Platen, Ma, Jernite, Plu, Xu, Scao, Gugger, Drame, Lhoest, and Rush}]{wolf-etal-2020-transformers}
Thomas Wolf, Lysandre Debut, Victor Sanh, Julien Chaumond, Clement Delangue, Anthony Moi, Pierric Cistac, Tim Rault, Rémi Louf, Morgan Funtowicz, Joe Davison, Sam Shleifer, Patrick von Platen, Clara Ma, Yacine Jernite, Julien Plu, Canwen Xu, Teven~Le Scao, Sylvain Gugger, Mariama Drame, Quentin Lhoest, and Alexander~M. Rush. 2020.
\newblock \href {https://www.aclweb.org/anthology/2020.emnlp-demos.6} {Transformers: State-of-the-art natural language processing}.
\newblock In \emph{Proceedings of the 2020 Conference on Empirical Methods in Natural Language Processing: System Demonstrations}, pages 38--45, Online. Association for Computational Linguistics.

\bibitem[{Wu et~al.(2019)Wu, Madotto, Hosseini-Asl, Xiong, Socher, and Fung}]{wu-etal-2019-transferable}
Chien-Sheng Wu, Andrea Madotto, Ehsan Hosseini-Asl, Caiming Xiong, Richard Socher, and Pascale Fung. 2019.
\newblock \href {https://doi.org/10.18653/v1/P19-1078} {Transferable multi-domain state generator for task-oriented dialogue systems}.
\newblock In \emph{Proceedings of the 57th Annual Meeting of the Association for Computational Linguistics}, pages 808--819, Florence, Italy. Association for Computational Linguistics.

\bibitem[{Yang et~al.(2012)Yang, Lai, and Lin}]{yang2012robust}
Miin-Shen Yang, Chien-Yo Lai, and Chih-Ying Lin. 2012.
\newblock A robust em clustering algorithm for gaussian mixture models.
\newblock \emph{Pattern Recognition}, 45(11):3950--3961.

\bibitem[{Young et~al.(2010)Young, Ga{\v{s}}i{\'c}, Keizer, Mairesse, Schatzmann, Thomson, and Yu}]{young2010hidden}
Steve Young, Milica Ga{\v{s}}i{\'c}, Simon Keizer, Fran{\c{c}}ois Mairesse, Jost Schatzmann, Blaise Thomson, and Kai Yu. 2010.
\newblock The hidden information state model: A practical framework for pomdp-based spoken dialogue management.
\newblock \emph{Computer Speech \& Language}, 24(2):150--174.

\bibitem[{Zhang et~al.(2020)Zhang, Hashimoto, Wu, Wang, Yu, Socher, and Xiong}]{zhang-etal-2020-find}
Jianguo Zhang, Kazuma Hashimoto, Chien-Sheng Wu, Yao Wang, Philip Yu, Richard Socher, and Caiming Xiong. 2020.
\newblock \href {https://aclanthology.org/2020.starsem-1.17} {Find or classify? dual strategy for slot-value predictions on multi-domain dialog state tracking}.
\newblock In \emph{Proceedings of the Ninth Joint Conference on Lexical and Computational Semantics}, pages 154--167, Barcelona, Spain (Online). Association for Computational Linguistics.

\bibitem[{Zhang et~al.(1996)Zhang, Ramakrishnan, and Livny}]{zhang1996birch}
Tian Zhang, Raghu Ramakrishnan, and Miron Livny. 1996.
\newblock Birch: an efficient data clustering method for very large databases.
\newblock \emph{ACM sigmod record}, 25(2):103--114.

\bibitem[{Zhu et~al.(2022)Zhu, Li, Mi, Zhu, and Huang}]{zhu-etal-2022-continual}
Qi~Zhu, Bing Li, Fei Mi, Xiaoyan Zhu, and Minlie Huang. 2022.
\newblock \href {https://doi.org/10.18653/v1/2022.acl-long.80} {Continual prompt tuning for dialog state tracking}.
\newblock In \emph{Proceedings of the 60th Annual Meeting of the Association for Computational Linguistics (Volume 1: Long Papers)}, pages 1124--1137, Dublin, Ireland. Association for Computational Linguistics.

\end{thebibliography}


\end{document}